\documentclass[twocolumn, aps, superscriptaddress, groupedaddress,nofootinbib]{revtex4} 

\usepackage{amsmath,amsthm,amsfonts,amssymb}  
\usepackage{algorithm}
\usepackage[noend]{algpseudocode}
\newtheorem{definition}{Definition}
\usepackage[english]{babel}
\usepackage[shortlabels]{enumitem}
\usepackage{graphicx,subfigure,multirow}
\usepackage{algorithm}
\usepackage[noend]{algpseudocode}

\usepackage[toc, page]{appendix}
 
\usepackage{lipsum}
\newtheorem{aid}{Intuition Aid}  
\newtheorem*{heuristic}{Description of heuristic method}
\newtheorem{open}{Open Problem}

\begin{document}  

\title{Leveraging directed causal discovery to detect latent common causes}
\author{Ciar{\'a}n M. Gilligan-Lee}
\affiliation{Spotify, London, United Kingdom\footnote{This work was started when CMGL was at Babylon Health}}
\affiliation{University College London, United Kingdom}
\author{Christopher Hart}
\affiliation{Babylon Health, London, United Kingdom}
\author{Jonathan G. Richens}
\affiliation{Babylon Health, London, United Kingdom}
\author{Saurabh Johri}
\affiliation{Babylon Health, London, United Kingdom}

\begin{abstract}
The discovery of causal relationships is a fundamental problem in science and medicine. In recent years, many elegant approaches to discovering causal relationships between two variables from observational data have been proposed. However, most of these deal only with purely directed causal relationships and cannot detect latent common causes. Here, we devise a general heuristic which takes a causal discovery algorithm that can only distinguish purely directed causal relations and modifies it to also detect latent common causes. We apply our method to two directed causal discovery algorithms, the Information Geometric Causal Inference of (Daniusis et al., 2010) and the Kernel Conditional Deviance for Causal Inference of (Mitrovic, Sejdinovic, \& Teh, 2018), and extensively test on synthetic data---detecting latent common causes in additive, multiplicative and complex noise regimes---and on real data, where we are able to detect known common causes. In addition to detecting latent common causes, our experiments demonstrate that both the modified algorithms preserve the performance of the original in distinguishing directed causal relations. 

\end{abstract}

\maketitle

\section{Introduction}

Causal knowledge is crucial in medicine, as causal relations---unlike correlations---allow one to reason about the consequences of possible treatments \citep{P,richens2019,perov2019}. Indeed, determining whether a particular treatment \emph{causes} a reduction in the severity of a disease is an important problem in healthcare. However, learning from medical records that a treatment is correlated with recovery is not sufficient to conclude that treatment will cause a new patient to recover, due to the presence of latent confounding. More expensive treatment could be given to wealthier patient's who are more likely to recover regardless of treatment, due to better lifestyle and diet. Here, the patients wealth acts as a latent confounder, or common cause, between receiving treatment and recovering. In reality, the treatment could be detrimental. Thus, learning the causal structure is vital. This problem is made harder when no controlled trial data exists----meaning the standard approach to learning causal structure isn't available. Hence, methods for discovering causal structure from uncontrolled, or observational, data are paramount to making correct treatment decisions.

Recently, a number of approaches to discovering causal relations between two variables have been proposed, including  \citep{Noise,shimizu2006linear,janzing2012information,mitrovic2018causal} among many others, which we describe in section~\ref{Previous approaches to causal discovery}. However, apart from \citep{Confounders} and \citep{lopez2015towards}, these deal only with purely directed causal relationships and cannot detect latent common causes. That is, given two variables $A$ and $B$, these algorithms can only distinguish $A$ causes $B$, graphically depicted in Figure~\ref{bivariate_cases}(a), from $B$ causes $A$, depicted in Figure~\ref{bivariate_cases}(b). Those algorithms that can detect latent common causes either impose stern restrictions on the underlying causal model---such as enforcing linearity \citep{shimizu2006linear,kaltenpoth2019we}, demanding that noise be additive \citep{Confounders}, or requiring huge amounts of training data \citep{lopez2015towards}. 
 As the presence of latent common causes is a significant problem, surmounting these shortcomings and developing methods for detecting latent common causes that do not rely on strong assumptions, such as additive noise or linearity, is an important problem in causal discovery.

Here, we devise a heuristic method which takes a purely directed causal discovery algorithm and modifies it so that it also discovers latent common causes. That is, our method takes as input a causal discovery algorithm which can only distinguish the causal structures in Figure~\ref{bivariate_cases}(a) and Figure~\ref{bivariate_cases}(b), and outputs a causal discovery algorithm which can distinguish between all three structures in Figure~\ref{bivariate_cases}. We apply our method to two directed causal discovery algorithms, Information Geometric Causal Inference (IGCI) algorithm of \citep{daniusis2012inferring} and  Kernel Conditional Deviance for Causal Inference (KCDC) algorithm of \citep{mitrovic2018causal}, which were both shown in \citep{mitrovic2018causal} to achieve state-of-the-art performance on the T{\"u}bingen cause-effect pair dataset\footnote{https://webdav.tuebingen.mpg.de/cause-effect/} \citep{mooij2016distinguishing}, with ICGI scoring 74\% and KCDC scoring 73\%. We extensively test on synthetic data---detecting latent common causes in additive, multiplicative and complex noise regimes---as well as on real data, where we are able to detect known common causes. Importantly, and in addition to detecting latent common causes with high accuracy, our experiments show that both modified algorithms don't sacrifice the performance of the original in distinguishing directed causal relations. 

\section{Related work} \label{Previous approaches to causal discovery}

In this work, causal structure will be graphically represented as a directed acyclic graph (DAG).
Methods for discovering the causal structure underlying a data generating process largely fall into two categories. The first, which we term \emph{global} causal discovery, attempts to reconstruct a (partially) undirected version of the causal structure. This approach is broadly split into two categories: constraint-based and score-based. The constraint-based approach employs conditional independence tests between the variables in question to determine which should share an edge in the causal structure. Examples include the PC algorithm and the FCI algorithm~\citep{Sprite}, the IC algorithm~\citep{P}, as well as algorithms which allow for latent variables \citep{silva2006learning} and selection bias \citep{spirtes1995causal}. 
The score-based approach introduces a scoring function, such as Minimum Description Length, that evaluates each network with respect to some training data, and searches for the best network according to this function \citep{friedman1997bayesian}. Greedy equivalent search (GES) \citep{chickering2002optimal} is a prime example of the score based causal discovery.
Hybrid approaches employing both constraint- and score-based techniques to outperform either alone \citep{tsamardinos2006max}. 

\begin{figure}[t] 
\centering
\begin{subfigure}
\centering
(a)\includegraphics[scale=0.21]{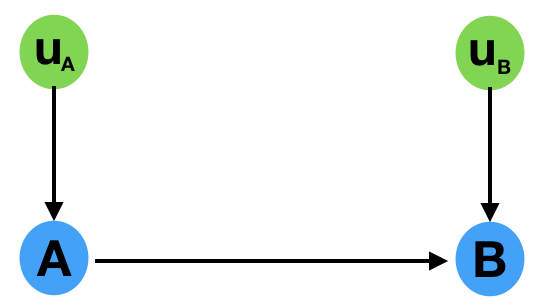}
\end{subfigure}
\quad
\begin{subfigure}
\centering
(b) \includegraphics[scale=0.21]{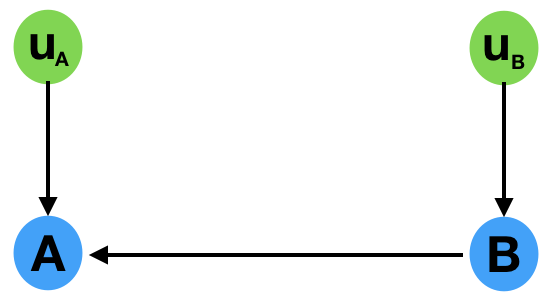}
\end{subfigure}
\quad
\begin{subfigure}
\centering
(c) \includegraphics[scale=0.21]{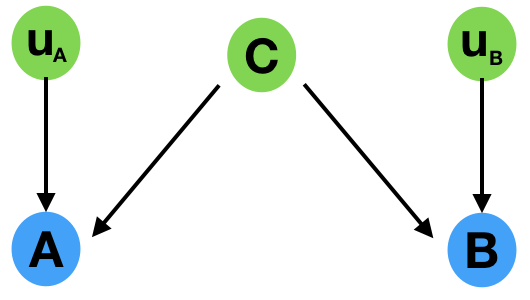}
\end{subfigure}
\caption{
Blue nodes represent observed variables, green nodes latent noise terms. (a) $A$ causes $B$. (c) $B$ causes $A$. (d) Latent common cause between $A$\&$B$.
}
\label{bivariate_cases}
\end{figure}

The main drawback of constraint-based algorithms is that, as they can only recover Markov equivalence classes, they are not always able to orient edges between dependent variables. Given correlated variables $A$ and $B$, these methods are unable to distinguish the structures depicted in Figure~\ref{bivariate_cases}. The second category of causal discovery algorithm, which we term \emph{local} or \emph{bivariate} causal discovery, aims to address this by exploiting notions of asymmetry between cause and effect. These methods specify different assumptions that aim to make such asymmetries testable at an observational level. 

The first type of assumption proposed in the literature, known as the functional causal model approach, specifies that each effect is a deterministic function of its cause together with some latent, independent noise term \citep{peters2017elements}. The first such algorithm, termed Linear Non-Gaussian Additive noise Model (LiNGAM) \citep{shimizu2006linear}, assumes the functions are linear and the latent noise variables are non-Gaussian. Given these assumptions, this method can distinguish between the causal structures in Figure~\ref{bivariate_cases}(a) and (b); a separate hidden variables extension~\citep{hoyer2008lingam} extends this result to all structures in Figure~\ref{bivariate_cases}. The other prototypical example, the Additive Noise Model (ANM) \citep{Noise}, allows the effect to be an arbitrary function of the cause, but assumes the effect only depends on the latent noise term additively. The ANM algorithm can distinguish between the two structures in Figure~\ref{bivariate_cases}(a)-(b) as long as the additivity assumption holds. ANM has been extended to allow the effect to depend on the cause and noise in a non-linear fashion \citep{zhang2009identifiability}. Finally, the Confounding Additive Noise (CAN) algorithm \citep{Confounders} extends the ANM algorithm to deal with latent common causes, allowing all structures in Figure~\ref{bivariate_cases} to be distinguished. We review the CAN model in Section~\ref{section: CAN model}.

The second type of assumption stipulates that the cause $P(\text{cause})$ be independent of the mechanism $P(\text{effect}|\text{cause})$. An information geometric approach to measuring such independence has been proposed \citep{daniusis2012inferring,janzing2012information}, known as the Information Geometric Causal Inference (IGCI) algorithm. This method can only distinguish the two structures in Figure~\ref{bivariate_cases}(a)-(b). 
IGCI does not require specific parametric functional relationships to hold. 

The last type of assumption entails that the causal mechanism $P(\text{effect}|\text{cause})$ should be ``simpler,'' given some quantifiable notion of ``simple'', than the anti-causal mechanism $P(\text{cause}|\text{effect})$. The Kernel Conditional Deviance Causal discovery (KCDC) algorithm \citep{mitrovic2018causal} algorithm uses conditional kernel mean embeddings to establish a norm on conditional probabilities and uses the variance of these as a simplicity measure. As experimentally shown in \citep{mitrovic2018causal}, IGCI and KCDC algorithms constitute the current state-of-the-art at distinguishing between the two causal structures in Figure~\ref{bivariate_cases}(a)-(b). 


While the ANM algorithm has been extended to deal with latent common causes via the CAN algorithm, its generalisations, such as KCDC and IGCI, have not. The current paper introduces a heuristic method that takes as input any purely directed causal discovery algorithm and outputs a new algorithm that can distinguish all causal structures in Figure~\ref{bivariate_cases}.

\subsection{Identifying latent common causes using additive noise models} \label{section: CAN model}

As a prelude to our method for detecting latent common causes, we review a current state-of-the-art algorithm for distinguishing all structures in Figure~\ref{bivariate_cases}, the Confounding Additive Noise model (CAN) of \citep{Confounders}. In CAN, one assumes that the functions underlying the causal structure from Figure~\ref{bivariate_cases}(c) are $A=f(C)+u_A$ and $B=g(C)+u_B$, for arbitrary $f,g$. The first step of CAN is to fit such a model using samples from $A,B$. To do this, manifold learning is employed to learn a representation of $C$ from the observed samples of $A,B$. Next, functions are fitted to map the learned representation of $C$ to $A$ and $B$ using e.g. Gaussian Process regression. The residuals resulting from this regression correspond to representations of the noise terms, $u_A$ and $u_B$. In fitting this model, one must ensure that $C, u_A, u_B$ are all mutually independent, as implied by the graphical structure of Figure~\ref{bivariate_cases}(c). To ensure independence, CAN employs a computationally expensive optimisation procedure. See Section~\ref{section:experiments} for further discussion.  

With such a model fit using samples of $A,B$, the next step is to use it to distinguish all structures in Figure~\ref{bivariate_cases}. Note that the additive noise assumption for Figure~\ref{bivariate_cases}(a)\footnote{Figure~\ref{bivariate_cases}(b) follows by interchanging $A$ and $B$.} means $B=g(A)+u_B$ and $A=u_A$, for arbitrary $g$. CAN distinguishes Figure~\ref{bivariate_cases}(a) from Figure~\ref{bivariate_cases}(c) if the variance of the learned $u_A$ is small, relative to the variance of $u_B$, and prefers Figure~\ref{bivariate_cases}(c) if the variances are approximately equal. The case of Figure~\ref{bivariate_cases}(b) follows by switching $A$ and $B$. The authors justify this as follows. Suppose $A$ causes $B$, and, by a slight measurement error, we observe $A'$, which differs from $A$ by a small additive term. Here, we would expect the functions to be as described above for Figure~\ref{bivariate_cases}(c). But we should not distinguish $A$ from its measurement result $A'$ if both variables almost coincide. In this case, after normalising $A,B$ to unit variance, the variance of $u_A$ would be small compared to $u_B$, as $A$ is close to $A'$. In general, the CAN algorithm distinguishes between all structures in Figure~\ref{bivariate_cases} using the following decision criteria: i) $A\rightarrow B$ if $\text{var}(u_A)/\text{var}(u_B)\ll 1$, ii) $A\leftarrow B$ if $\text{var}(u_A)/\text{var}(u_B) \gg1$, and iii) $A\leftarrow C \rightarrow B$ if $\text{var}(u_A)/\text{var}(u_B)\approx 1$. Determining thresholds for the above criteria is a tricky problem, with values being decided on a case by case basis; see \citep{Confounders} for more information. 




\section{Methods} \label{section: reduction to common causes and directed causal discovery}




We work in the functional causal model framework \citep{peters2017elements}. Here, a causal structure corresponds to a directed acyclic graph (DAG) between observed and latent variables, with each variable a deterministic function of its parents and some latent, independent noise term. For instance, in Fig.~\ref{bivariate_cases}(a), this means $A=f(u_A)$ and $B=g(A,u_B)$ for deterministic functions $f,g$, while in Fig.~\ref{bivariate_cases}(c) it means $A=f(C,u_A)$ and $B=g(C,u_B)$. A causal model corresponds to a DAG, the functions, and a specification of the priors over the latent noise terms. Hence one can write the observed probability of $A=a$ given $B=b$ as $P(A=a|=b) = \sum_{u| a=f(b, u)} Q(U=u)$ where $f$ is the deterministic function mapping $B$ and the latent noise term $U$ to $A$, and $Q(U)$ is the prior over the latent noise term. We will say that two causal models are \emph{observationally equivalent} if they are characterised by the same set of distributions over observed variables. Denote the set of distributions over observed variables, which is specified by the functions and latent priors as discussed above, for a particular causal model $M$, by $P_M$. Two causal models $M$ and $M'$ are then \emph{observational equivalent} if and only if $P_M=P_M'$.

\subsection{Description and intuition of our heuristic algorithm}\label{section: description of heuristic methid}

In this section we will describe our heuristic and provide an outline of the intuition it is derived from. First, we will provide a quick description of our method, the mechanics of which will then be explained in the rest of the section. 

\begin{heuristic}
We start by fitting a curve to samples of observations $A_i, B_i$ (using a manifold learning algorithm such as isomap \citep{tenenbaum2000global}, say) which yields a one-dimensional parametrisation $T_i$ of the data. Proceeding, the heuristic is motivated by the following intuition:
\begin{enumerate}
\item If $A \leftarrow C \rightarrow B$ then $T$ is expected to essentially correspond to $C$ and directed
causal discovery algorithms applied to $(A, T)$ and $(B, T)$ are expected to
conclude that arrows are outgoing from $T$.
\item If $A \rightarrow B$ ($A \leftarrow B$ follows similarly), then---as we'll justify in this section---$T$ is expected to essentially correspond to $f(u_A)$, where $A = f(u_A)$, and directed causal discovery algorithms applied to $(A, T)$ yield inconclusive and ambiguous results, while for $(B, T)$ they output a causal arrow outgoing from $T$ (as $B = g(A, u_B) = g(f(T), u_B)$ as we'll discuss in this section.
\end{enumerate}
This asymmetry can then be used to distinguish $A \rightarrow B$, $A \leftarrow B$ and $A \leftarrow C \rightarrow B$.
\end{heuristic}

We will now proceed with explaining why each of the above steps follows. To this end, and to help with exposition, we have broken down the required conceptual components and present them as \emph{Intuition Aids}. 

We start by defining a causal model with no directed arrows between observed variables as \emph{purely common cause}. 

\begin{aid}\label{lemma 1}
Any causal model with a directed arrow between $A$ \& $B$ ($A\rightarrow B$ or $B \rightarrow A$) is observationally equivalent\footnote{See start of Section~\ref{section: reduction to common causes and directed causal discovery} for the definition of observational equivalence.} to one that is purely common cause.
\end{aid}

\begin{figure}[t] 
\centering
\includegraphics[scale=0.22]{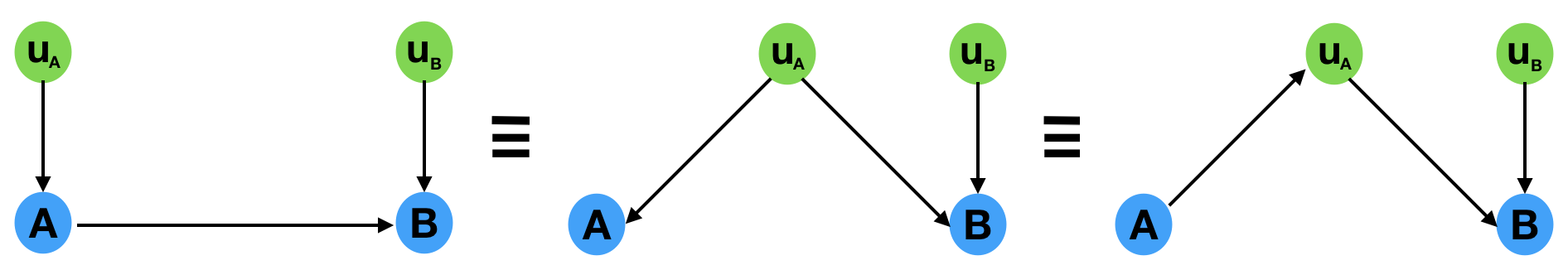}
\caption{Depiction of three observationally equivalent DAGs discussed in Intuition Aid~\ref{lemma 2}. The symbol $\equiv$ denotes observational equivalence. 
}
\label{direct_to_common_cases}
\end{figure}

This result has been shown before, in~\citep{lee2017causal} for example, but to see that it holds for the case of Figure~\ref{bivariate_cases}(a), consider the following. A model for this causal structure corresponds to specifying the functional dependencies between $A,B$ and their parents: $A=f(u_A)$ and $B=g(A,u_B)$. Substituting $f(u_A)$ for $A$ in $B$, one obtains $B=g(f(u_A), u_B)$. Defining $u'_A:= f(u_A)$ results in the relations $A=u'_A$ and $B=g(u'_A,u_B)$. The directed arrow from $A$ to $B$ has been replaced by the latent ``common cause'' $u_A'$, and Figure~\ref{bivariate_cases}(a) has been reduced to a purely common cause model. This is outlined graphically in the first two causal structures from Figure~\ref{direct_to_common_cases}, where $\equiv$ denotes observational equivalence. 

We say a causal model is in \emph{canonical form}\footnote{Note that this notion of canonical model is different to the one discussed in \citep{hoyer2008lingam}. There a model is canonical if all functions are linear and each latent variable is a root node with at least two children. Furthermore, although different latent variables may have the same sets of children, no two latent variables should exhibit exactly proportional sets of connection strengths to the observed
variables.} when it is purely common cause. Note that Figure~\ref{bivariate_cases}(c) is already in canonical form. The following Intuition Aid provides another key observation underlying our approach.

\begin{aid} \label{lemma 2}
Given the causal model in Figure~\ref{bivariate_cases}(a)\footnote{The case of Figure~\ref{bivariate_cases}(b) follows by interchanging $A$ and $B$.}, the canonical model is consistent with two causal structures: one where $u_A$ is a common cause between $A$ and $B$, and the other where $u_A$ is the mediator of the causal arrow from $A$ to $B$. This is depicted in the last two structures of  Figure~\ref{direct_to_common_cases}.
\end{aid}

Indeed, without loss of generality, the canonical model of Figure~\ref{bivariate_cases}(a) is given by $A=u_A$ and $B=g(u_A,u_B)$. These functions are consistent with the second causal structure in Figure~\ref{direct_to_common_cases}, in which a causal arrow points from $u_A$ to $A$. But the equality $A=u_A$ is also consistent with the third causal structure in Figure~\ref{direct_to_common_cases}, where the causal arrow points from $A$ to $u_A$. That is, the relationship $A=u_A$ is consistent with both an arrow from $u_A$ to $A$ and an arrow from $A$ to $u_A$. Note that due to the independence of the noise terms $u_A$ and $u_B$, this is not true of the other arrows in the causal structure. A causal arrow pointing from $B$ to $u_A$ in Figure~\ref{direct_to_common_cases} would induce correlations between $u_A$ and $u_B$, in direct contradiction with the independence of noise terms.

Intuition Aid~\ref{lemma 2} tells us that in the canonical model of Figure~\ref{bivariate_cases}(a), the directionality of the causal link between $A$ and $u_A$ is underdetermined, as illustrated in the last two structures of  Figure~\ref{direct_to_common_cases}. Thus applying a directed causal discovery algorithm to $A$ and $u_A$ will result in inconclusive and ambiguous results. All other causal links are fully determined and directed however. If, on the other hand, the original structure had been Figure~\ref{bivariate_cases}(b), then the causal link between the canonical ``common cause'', $u_B$, and $B$ would be underdetermined, with all remaining links determined and directed. Note that if the structure had originally been that of Figure~\ref{bivariate_cases}(c), then neither arrow from the true common cause, $C$, to the observed variables is underdetermined. 

This observation is what underlies the conclusions from items (a) and (b) outlined at the start of this section. In the next two subsections we will describe how to implement our method.




\subsection{From intuition to implementation} \label{section: detecting common causes}

The presentation of our heuristic method has thus far has not formally defined the notion of an underdetermined causal relation. We now present this definition. Here we focus on directed causal discovery algorithms, $D$, that decide the causal direction using a quantifiable notion of asymmetry, such as IGCI \citep{daniusis2012inferring} and KCDC \citep{mitrovic2018causal}, described in Section~\ref{Previous approaches to causal discovery}. In general, such algorithms assign a real scalar to each causal direction $v_{A\rightarrow B}^D$ and $v_{B\rightarrow A}^D$, returning the causal direction with smallest $v$ value. Due to the pervasiveness of statistical noise, such algorithms have a decision threshold $\delta$, such that if $\vert v_{A\rightarrow B}^D - v_{B\rightarrow A}^D \vert < \delta$ the algorithm fails to detect a causal direction.

As an illustration, the KCDC algorithm of \citep{mitrovic2018causal} uses conditional kernel mean embeddings to establish a norm on conditional probabilities $P(B|A=a)$, $P(A|B=b)$ for each value of $A=a$, $B=b$. If the variance $v_{A\rightarrow B}$ of the norms $\{\Vert P(B|A=a) \Vert\}_a$ is smaller than the variance $v_{B\rightarrow A}$ of $\{\Vert P(A|B=b) \Vert\}_b$, the KCDC algorithm returns Figure~\ref{bivariate_cases}(a), otherwise it returns Figure~\ref{bivariate_cases}(b). 

The following definition formalises the notion of a causal relation being underdetermined relative to a particular directed causal discovery algorithm.
\begin{definition} \label{def: under}
A causal relation between two variables $A$ and $B$ is \emph{underdetermined} relative to causal discovery algorithm $D$ if $\vert v_{A\rightarrow B}^D - v_{B\rightarrow A}^D \vert < \delta$ for pre-specified decision threshold $\delta$.
\end{definition}


Finally, to implement our heuristic method, the canonical common cause needs to be learnt from samples of the observed variables $A,B$. At the start of Section~\ref{section: description of heuristic methid} we stated that manifold learning or dimensionality reduction algorithms, such as the isomap algorithm of \citep{tenenbaum2000global}, would perform this requirement. Indeed, under the hypothesis that such an algorithm will learn the most likely latent representation of the data, this seems intuitively to be the case. However, for this intuition to be borne out,  we need to stipulate that the influence of the latent noise terms on the observed variables be small. That is, we need to require that fluctuations around the manifold parameterised by the canonical common cause due to latent noises be small for the canonical common cause from samples of the observed variables $A,B$. 

\begin{aid} \label{lemma 3}
Under the assumption that the influence of the latent noise term is small, the canonical common cause can be identified from samples of $A,B$.
\end{aid} 

\begin{proof}
Consider the causal model of Figure~\ref{bivariate_cases}(a)\footnote{All other cases follow by a similar argument.}, which, following Lemma~\ref{lemma 1} we can write as $A=u_A$ and $B=g(u_A, u_B).$ Now, a Taylor expansion yields $B=g(u_A,0) + g'(u_A,0)u_B \,+ \text{ (higher order terms in } u_B) $, where $g'$ is the partial derivative of $g$ with respect to $u_B$. The assumption that the influence of the noise term $u_B$ on $B$ is small corresponds to the fact that we can drop higher order terms in $u_B$ in the above Taylor expansion. Hence, we can write $B=g(u_A,0) + g'(u_A,0)u_B$. Without loss of generality, we can assume the noise term is a Gaussian random variable with mean $0$. This implies that the expected value of $B$ is $B=g(u_A,0)$. Thus the expected value of $(A,B)$ samples is $(u_A, g(u_A,0))$. That is, they are both functions of the canonical common cause $u_A$ alone and can hence be identified from $A,B$ samples.
\end{proof}

This implies that given the canonical causal model associated with any of the structures from Figure~\ref{bivariate_cases}, and the assumption of small influence of noise, manifold learning can be utilised---as is done in the CAN algorithm \citep[Section 3]{janzing2012information}, described in Section~\ref{section: CAN model}---to determine the approximate parameterisation (up to rescaling) of the canonical ``common cause'' from samples of the observed $A,B$. As discussed at the start of this section, one can use this to determine the causal structure underlying $A$ and $B$ as follows.

In the case of Figure~\ref{bivariate_cases}(a) and Figure~\ref{bivariate_cases}(b), the ``common cause'' corresponds to the variables $u_A$ and $u_B$ respectively. In the case of Figure~\ref{bivariate_cases}(c) it is the actual common cause, $C$. Second, given the parameterisation of the ``common cause,', one can determine the original pre-canonical causal structure using a purely directed causal discovery algorithm to determine which causal link to the observed variables is underdetermined (if any). If the arrow to $A$ is underdetermined then the causal structure is Figure~\ref{bivariate_cases}(a); if the arrow to $B$ is underdetermined the structure is Figure~\ref{bivariate_cases}(b); if no arrow is underdetermined, the structure is Figure~\ref{bivariate_cases}(c).

Our algorithm is described below in Algorithm~\ref{causal_discovery}.

\begin{algorithm}[]
\caption{
}\label{causal_discovery}
\textbf{Input:} $A,B$ data, manifold learning algorithm $M$, directed causal discovery algorithm $D$. 
\\
\textbf{Output:} Single causal structure from Figure~\ref{bivariate_cases}.
\begin{algorithmic}[1]
\State Run $M$ on $\{A,B\}$ to obtain parameterisation (up to rescaling, etc.) of canonical common cause $T$ that best fits the data. 
\State Implement $D$ between $T$\&$A$ and $T$\&$B$ \\
\textbf{if} $D$ outputs $T\rightarrow B$ \& the causal link between $T$\&$A$ is underdetermined relative to $D$ \textbf{do:} \\
\qquad Output DAG from Figure~\ref{bivariate_cases}(a) \\
\textbf{else if} $D$ outputs $T\rightarrow A$ \& the causal link between $T$\&$B$ is underdetermined relative to $D$ \textbf{do:} \\
\qquad Output DAG from Figure~\ref{bivariate_cases}(b) \\
\textbf{else if} $D$ outputs $T\rightarrow B$ \& $T\rightarrow A$ \textbf{do:} \\
\qquad Conclude $T$ is a common cause of $A,B$ and output DAG from Figure~\ref{bivariate_cases}(c) \\
\textbf{return} DAG output from above
\end{algorithmic}
\end{algorithm} 

The experiments performed in Section~\ref{section:experiments} show empirically that Algorithm~\ref{causal_discovery} performs well even on synthetic data when the influence of latent noise is not small. This suggests that the canonical common cause can be learned from $A,B$ in a wider array of settings. In Section~\ref{identify} we set out an open question that would allow the canonical common cause to be learned from $A,B$ samples under weaker assumptions than that discussed above. 

\subsection{Decision criteria} \label{section: decision criteria}

Due to the pervasiveness of statistical noise, we now introduce a heuristic decision criterion for checking the conditions of Algorithm~\ref{causal_discovery}, which is inspired by the decision threshold for KCDC and IGCI discussed in the last section. Consider step $3$ of Algorithm~\ref{causal_discovery}, where the learned canonical common cause is denoted $T$. To simultaneously check whether the causal link between $A$ and $T$ is underdetermined, and also whether $T\rightarrow B$, are consistent with the data, one could apply the following heuristic. Check if 
$$ \Delta = \frac{\bigg| \vert v_{A\rightarrow T} - v_{T\rightarrow A}   \vert - \vert  v_{B\rightarrow T} - v_{T\rightarrow B}  \vert \bigg| }{\max\left( \vert v_{A\rightarrow T} - v_{T\rightarrow A} \vert, \vert  v_{B\rightarrow T} - v_{T\rightarrow B} \vert \right)} 
$$
lies in the region $(\alpha, 1]$ for some pre-specified decision threshold $\alpha$ near $1$. This follows as the conditions in step $3$ imply $\vert v_{A\rightarrow T} - v_{T\rightarrow A}   \vert< \delta$ and $\vert  v_{B\rightarrow T} - v_{T\rightarrow B}  \vert >\delta$, for small $\delta$. Thus, the first term in the numerator is generally close to zero, while the second is not.

Alternatively, if $\Delta \in (0, \alpha]$, then the data is consistent with $T\rightarrow B$ \& $T\rightarrow A$, as in this case $\vert v_{A\rightarrow T} - v_{T\rightarrow A}   \vert > \delta$ and $\vert  v_{B\rightarrow T} - v_{T\rightarrow B}  \vert >\delta$. Hence, both terms in the numerator can be roughly thought of as being closer in magnitude than in the case of steps 3 and 5. This allows us to check whether step 7 of Algorithm\ref{causal_discovery} is true.

Note that the first criterion $\Delta \in (\alpha, 1]$ is also consistent with step 5 of Algorithm~\ref{causal_discovery}, as steps 3 and 5 are the same under interchange of $A$ \& $B$. Hence if $\Delta \in (\alpha, 1]$, then either step 3 or step 5 is true, and step 7 is false, as per the above. As the two cases in in step 3 and 5 are purely directed causal structures, they can be distinguished using the original directed causal discovery algorithm.

Note that the above discussion is heuristic in nature, $\Delta$ is not mathematically constrained to lie in the specific regions described. However, based on the theoretical intuition developed in the last Section, it seems a reasonable heuristic to adopt. In the experiments described in Section~\ref{section:experiments}, this heuristic performs very well on both synthetic and real data.

To assign confidence to the outcome of the above heuristic, one could take a bootstrapped approach and calculate the mean of the $\{\Delta_i\}$'s output by running the algorithm on subsamples of the input data, where $\Delta_i$ is the value computed on the $i$th subsample. We suppress the index $i$ in the below for notational ease. As reasoned above, $\text{mean}(\{\Delta\})\in (\alpha, 1]$ is consistent with a directed causal structure and $\text{mean}(\{\Delta\})\in (0, \alpha]$ with a common cause. Additionally, the variance of the $\{\Delta\}$'s calculated in this manner encodes information of the correct DAG. For instance, small variance $\text{var}(\{\Delta\})\in (0, \gamma]$---where $\gamma$ is again a pre-specified decision threshold less than $1$---is consistent with a directed causal structure and large variance $\text{var}(\{\Delta\})\in (\gamma, 1]$ is consistent with a common cause. This follows because for a directed causal structure only one of the terms in the numerator of each $\Delta_i$ must lie in the region $[0,\delta)$, for small $\delta$, and hence cannot vary much. For a common cause, both terms are outside this region and thus can vary more. 

These two implication, high $\text{mean}(\{\Delta\})$ indicating a directed cause and high $\text{var}(\{\Delta\})$ a common cause, can be used in conjunction to determine the causal structure. For example, it could be the case that a particular common cause structure has moderately high $\text{mean}(\{\Delta\})$, but such a high $\text{var}(\{\Delta\})$ that a directed cause is ruled out, and a common cause is returned by the algorithm. Similar reasoning can be applied to a directed cause whose $\text{mean}(\{\Delta\})$ is lower than expected, but has such a low $\text{var}(\{\Delta\})$ that a common cause is ruled out.

To summarise, the above decision criteria are as follows. The mean and variance of $\{\Delta\}$'s output by running the algorithm on subsamples of the input data are computed. Recall the possible range of $\text{mean}(\{\Delta\})$ is $[0,1]$. This range is split into a number of regions. We fix region $1$ to be $[0,\alpha_1]$, region $2$ to be $[\alpha_1, \alpha_2]$, and so on. Each region $R$ is associated with a threshold value for $\text{var}(\{\Delta\})$, $\gamma_R$. If $\text{mean}(\{\Delta\})$ lies in region $R$, then if $\text{var}(\{\Delta\})\leq \gamma_R$ the algorithm outputs a directed causal structure, whose direction is then determined by the original directed causal discovery algorithm. But if $\text{var}(\{\Delta\})> \gamma_R$, a common cause structure is output. Note that as the regions get closer to $0$, that is, to $[0,\alpha_1]$, the variance threshold $\gamma_R$ gets increasing small. Hence, $\gamma_R < \gamma_{R+1}$. This encodes our previous reasoning of how mean and variance imply different causal structures. Additionally, there should be a failure mode for very low variance and mean, and very high mean and variance, as these are not indicative of a direct nor a common cause. 


As with the decision criteria for the CAN algorithm, described in Section~\ref{section: CAN model}, and KCDC \citep{mitrovic2018causal}, setting such is a challenge, due to the varying impact of noise within experimental data sets. An appropriate method we take in this work is to use data of similar provenance with known causal relations and determine appropriate thresholds for that domain, which are then fixed.

\subsection{Open questions and future extensions} \label{identify}




Given the assumption that the strength of latent noise is relatively small, a representation of the canonical common cause can be extracted from samples of the observed $A,B$. The discussion in Section~\ref{section: detecting common causes} and Intuition Aids~\ref{lemma 1} \& \ref{lemma 2} show how this representation can be used to distinguish all three causal structures in Figure~\ref{bivariate_cases}. 

In the experiments performed in Section~\ref{section:experiments}, and specifically in Section~\ref{section: sensitivity}, it's shown empirically that while the performance of Algorithm~\ref{causal_discovery} does decrease when the influence of latent noise is not small, it is quite robust to the strength of latent noise. This gives hope that the canonical common cause can be learned from $A,B$ in a wider array of settings than was shown in this Section. We now set out an open question that would allow the canonical common cause to be learned from $A,B$ samples under weaker assumptions than that discussed above. 

\begin{open}
Given samples from $A=f(C, u)$ and $B=g(C, v)$, under what conditions can $C$ be learned?
\end{open}

Some progress has been made toward the solution of this open problem by \citep{gresele2020incomplete}, who show that given $A=f(C, u)$ and $B=g(C, v)$, $C$ can be extracted as long as $A$ and $B$ are ``sufficiently different,'' where, informally, this means the distribution of $u$ and $v$ should
vary significantly when conditioning on different values of the $C$.

\subsection{Future extensions: detecting simultaneous direct and latent common causes} \label{section: Detecting simultaneous direct and latent common causes}

\begin{figure}[t]  
\centering
\begin{subfigure}
\centering
(a)\includegraphics[scale=0.21]{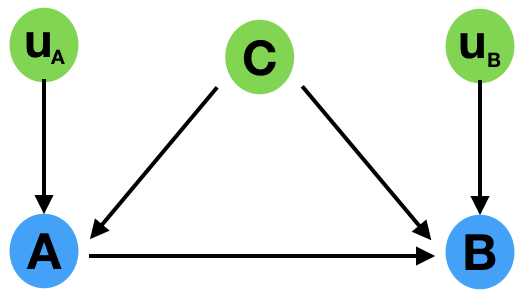} 
\end{subfigure}
\quad
\begin{subfigure}
\centering
(b) \includegraphics[scale=0.21]{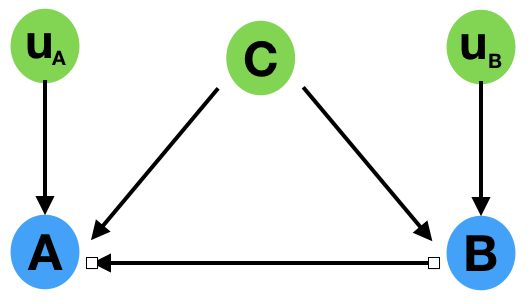}
\end{subfigure}
\begin{subfigure}
\centering
(c) \includegraphics[scale=0.2]{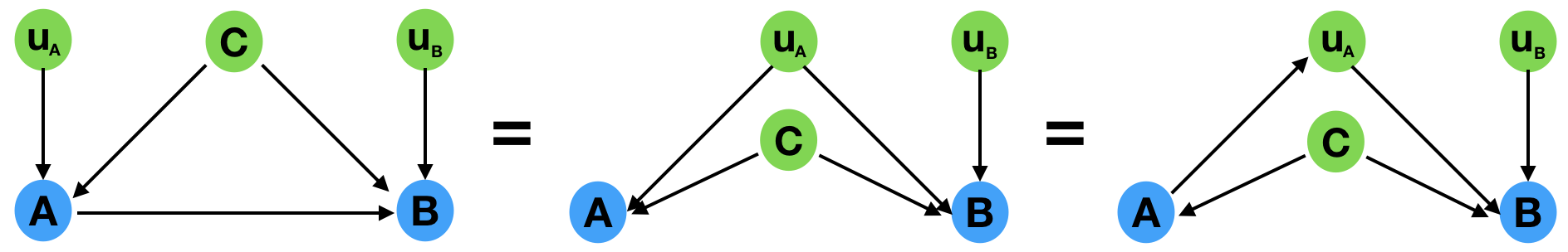}
\end{subfigure}
\caption{
Blue nodes represent observed variables, green latent noise terms. (a) $A$ causes $B$ with latent common cause. (b) $B$ causes $A$ with latent common cause. (c) Reduction from direct and common cause to two common causes.
}
\label{simultaneous_direct_and_common}
\end{figure}

We now discuss the extension of our Algorithm to the situation where there is both a directed cause and a latent common cause. To distinguish between the two causal structures in Figure~\ref{simultaneous_direct_and_common}, a similar procedure as Section~\ref{section: reduction to common causes and directed causal discovery} can be applied. In this case, the directed arrow between $A$ and $B$ can be mathematically replaced by another common cause, resulting in a purely common cause structure with two common causes. When determining if an arrow from one of the common causes can be reversed, to check which of the two structures from Figure~\ref{simultaneous_direct_and_common} is consistent with the data, all remaining common causes must be conditioned upon, or held fixed. The rational behind reversing an arrow is the same as that given in Section~\ref{section: reduction to common causes and directed causal discovery} and is graphically outlined in Figure~\ref{simultaneous_direct_and_common} (c). 

The difference between this case and the one presented in Section~\ref{section: reduction to common causes and directed causal discovery} is the the dimensionality reduction algorithm must here output the parameterisations (up to individual rescaling) of two common causes rather than one. However, for the above described reversing of arrows to work, the dimensionality reduction algorithm must output independent and individual parameterisations of each common cause. Hence, non-linear blind source separation \citep{jutten2003advances} or disentangling of latents \citep{disentangledlatenst2018} algorithms must be employed. This will effect the identifiability of the algorithm, as there are certain functional relationships for which blind source separation fails to return the correct parameterisation. This will be explored further in future work.

\section{Experimental results} \label{section:experiments}

We test our method by showing it can turn both the KCDC algorithm  of~\citep{mitrovic2018causal} and the IGCI algorithm of~\citep{daniusis2012inferring}, which can each only distinguish Figure~\ref{bivariate_cases}(a) and (b), into algorithms that can distinguish all causal structures in Figure~\ref{bivariate_cases}. We refer to the modified KCDC and IGCI algorithms output by our method as modKCDC and modIGCI respectively. To benchmark performance of modKCDC and modIGCI, we compare against CAN \citep{Confounders} the only previous causal discovery algorithm explicitly designed to distinguish all structures in Figure~\ref{bivariate_cases}, which we outlined in Section~\ref{section: CAN model}. The CAN algorithm proved to be very computational intensive to run compared to modKCDC and modIGCI. To reduce computational cost, we only enforced two of the three independence constraints discussed in Section~\ref{section: CAN model}: $u_A, u_B \perp C$.

For KCDC, we use a radial basis function kernel with hyperparameters fixed in all experiments; there is scope to extend this to multiple kernel majority vote following \citep{mitrovic2018causal}, though performance with a single kernel was effective here. We use the implementation of IGCI from \citep{daniusis2012inferring}. We use the implementation of CAN outlined in \citep{Confounders}, using Gaussian Process regression, HSIC to test independence, and employing up to five iterations of the independence optimisation loops, capped at 5000 iterations of the solver each time. For the manifold learning subroutine of modKCDC, modIGCI, and CAN, we employ Isomap~\citep{tenenbaum2000global}. This is the implementation of manifold learning originally used for CAN  \citep{Confounders}. All data was normalised to have mean 0 and variance 1.

We follow the bootsrapping procedure outlined in Section~\ref{section: decision criteria}, using 25 bootstraps each randomly sampling 95\% of the data. To set thresholds, we generated synthetic data from directed causal models with additive, multiplicative, and complex noise functions and determined the threshold values which approximately reproduced the performance of the original directed causal discovery algorithms on these different models. We did this for modKCDC and modIGCI, as well as for CAN, whose decision criteria were discussed in Section~\ref{section: CAN model}. 
This is similar to the way the CAN thresholds were set in \citep{janzing2012information}. We report the thresholds set in this manner in the Appendix. These thresholds are used in all experiments.

\subsection{Synthetic data} \label{section:synthetic}


We first test on synthetic data for both purely directed and purely common causes. The functions used to generate this synthetic data were adapted from the work of \citep{mitrovic2018causal}, who used them to test performance of KCDC and compare its performance to IGCI. As such, we believe using similar synthetic functions and the same range of synthetic noise distributions provides a fair and robust benchmark of KCDC, IGCI, and their modifications.  

On the synthetic directed datasets, we compare the modified algorithms to the original to determine whether the modification noticeably reduced the performance in detecting directed causal structure. Our experiments show modKCDC and modIGCI preserve the level of performance of the original KCDC and IGCI algorithms.  

\subsubsection{Directed causal structures} \label{section:directedexperiments}
In the below experiments, we sample 100 datasets each of 250 observations and test modKCDC and modIGCI. The CAN algorithm proved to be very computational intensive to run, taking 300-500 times as long to run on a dataset of size 250 compared to modKCDC and modIGCI. Hence we only tested it on 10 datasets, each of 250 observations. For our three algorithms, we sample $A\in \mathcal{N}(0,1)$, and test across three different noise regimes: Normal $n_B \in\mathcal{N}(0,1)$, Uniform $n_B \in\mathcal{U}(0,1)$, and Exponential $n_B \in\mathcal{E}(1)$. We record the accuracy over all datasets, that is, the percentage of cases where the algorithm output the correct ground truth directed structure. This is the same metric employed by \citep{mitrovic2018causal} in their experiments. The CAN algorithm was only able to fit a model that satisfied the independence criteria discussed in Section~\ref{section: CAN model} and~\ref{section:experiments} in the case of additive noise, as used in experiments (1) and (2) below. This is to be expected as non-additive noise, as in experiments (3)--(6), violates the identifiability requirements of CAN. In these cases we could not report CAN's performance. Results are presented in Table~1.

\vspace{1mm}

\noindent\textbf{Additive noise:} $$\begin{aligned}
(1) \text{ } B&=\sin{(10A)} + e^{3A} + n_B 
\\ 
(2) \text{ } B&= A\exp{A^2} + n_B
\end{aligned}
$$



 
 

\noindent\textbf{Multiplicative noise:} 
$$\begin{aligned} 
(3) \text{ } B&=\big(\sin{(10A)} + e^{3A}\big)e^{n_B} \\
(4) \text{ } B&=\big(A^2 + A^5\big)e^{n_B}
\end{aligned}
$$


 
 

\noindent{\textbf{Complex noise:}}
$$\begin{aligned}
(5) \text{ } B&=A^5 - \sin{(A^2 |n_B|)} \\
(6) \text{ } B&= \log{(A + 10)} + A^{8n_B}
\end{aligned}
$$

\begin{center}
\begin{tabular}{c}
 Table 1: Directed cause experiment results \\
 \begin{tabular}{lllrrr}
\hline
 \multicolumn{5}{c}{\textbf{Directed Cause}}\\
 Experiment & Algorithm & Normal &
 Uniform & Expon. \\
 \cline{1-5}
\multirow{ 5}{*}{1} & modKCDC & $99\%$ & $100\%$ &  $98\%$ \\
 & KCDC      & $100\%$    & $100\%$ &  $100\%$   \\
 & modIGCI & $100\%$ & $99\%$ &  $100\%$ \\
 & IGCI      & $100\%$    & $100\%$ &  $100\%$   \\
 & CAN      & $40\%$    & $33\%$ &  $57\%$   \\
  \cline{1-5}
\multirow{ 5}{*}{2} & modKCDC & $99\%$    & $100\%$ &  $98\%$ \\
 & KCDC      & $100\%$    & $100\%$ &  $100\%$   \\
 & modIGCI & $100\%$    & $100\%$ &  $100\%$ \\
 & IGCI      & $100\%$    & $100\%$ &  $100\%$   \\
 & CAN      & $43\%$    & $66\%$ &  $50\%$   \\
         \cline{1-5}
\multirow{ 4}{*}{3} & modKCDC      & $100\%$    & $100\%$ &  $100\%$ \\
 & KCDC      & $100\%$    & $100\%$ &  $100\%$   \\
 & modIGCI      & $100\%$    & $99\%$ &  $98\%$ \\
 & IGCI      & $100\%$    & $100\%$ &  $100\%$   \\
 \cline{1-5}
\multirow{ 4}{*}{4} & modKCDC      & $100\%$    & $99\%$ &  $100\%$ \\
 & KCDC      & $100\%$    & $100\%$ &  $100\%$   \\
 & modIGCI      & $99\%$    & $100\%$ &  $96\%$ \\
 & IGCI      & $100\%$    & $100\%$ &  $100\%$   \\
 \cline{1-5}
\multirow{ 5}{*}{5} & modKCDC      & $98\%$    & $98\%$ &  $98\%$ \\
 & KCDC      & $100\%$    & $100\%$ &  $100\%$   \\
 & modIGCI      & $99\%$    & $98\%$ &  $99\%$ \\
 & IGCI      & $100\%$    & $100\%$ &  $100\%$   \\
 \cline{1-5}
\multirow{ 4}{*}{6} & modKCDC   & $100\%$    & $97\%$ &  $99\%$ \\
 & KCDC      & $100\%$    & $100\%$ &  $100\%$   \\
 & modIGCI   & $29\%$    & $57\%$ &  $99\%$ \\
 & IGCI      & $31\%$    & $100\%$ &  $100\%$   \\
    \end{tabular}
    \end{tabular}
\end{center}

In all synthetic direct cause experiments, KCDC and IGCI perform extremely well, achieving 100\% accuracy in all situations with the exception of IGCI in experiment (6) with normally distributed noise. In all cases bar one, modKCDC and modIGCI either match this performance, or come within 1-2\% of it. It is also interesting to note that the one case in which IGCI performed badly, so did modIGCI. modIGCI failed to match the performance of IGCI in one case however, that of experiment (6) with uniform noise. As IGCI itself performed badly on this experiment with normal noise, it seems to present a challenging dataset for this algorithm, so it is perhaps not surprising that modIGCI yielded poor performance in this case. 


The CAN algorithm did not perform well on any of these synthetic datasets, failing to fit a model that complied with the independence criteria in all cases beyond additive noise. We only reported accuracy for CAN in cases where it successfully fit a model. This in principle gives CAN an added advantage. In (1), this happened in 5/10, 9/10, 7/10 cases for the three different noise distributions. In (2), this happened in 7/10, 3/10, 5/10 cases. CAN's poor performance shows the difficulty in developing an algorithm that can distinguish purely directed and common causes.

\subsubsection{Common cause}
The common cause setup closely follows that of Section~\ref{section:directedexperiments}. For these synthetic experiments, the common cause $T$ was sampled from $T\in \mathcal{N}(0,1)$. Here, CAN was only able to fit a model for experiment (2), below. In all other cases we could not report CAN's performance. Results are presented in Table~2.

\vspace{1mm}
\noindent{\textbf{Additive noise:} }
$$
\begin{aligned}
(1) \text{ } A &= \sin(10 T) + e^{3T} + n_A \\
B &= \log(T + 10) + T^6 + n_B \\
(2) \text{ }A &= \log(T + 10) + T^6 + n_A \\
 \text{ } B &= T^2 + T^6 + n_B.
\end{aligned}
$$



\noindent{\textbf{Multiplicative noise:} }
$$ \begin{aligned} 
(3) \text{ }  A&= \big(\sin(10 T) + e^{3T}\big)e^{n_A} \\
B&= \big(T^2 + T^6\big)e^{n_B} \\
(4) \text{ }   A&= \big(\sin(10 T) + e^{3T}\big)e^{n_A} \\
B&= \big(\log(T + 10) + T^6\big)e^{n_y}
\end{aligned}
 $$



\noindent{\textbf{Additive and Mulitplicative noise:}}
$$ \begin{aligned}
(5) \text{ } A&= \log(T + 10) + T^6 + n_A \\
B&= \big(T^2 + T^6\big)e^{n_B} \\
(6) \text{ } A&= \sin(10 T) + e^{3T} + n_A \\ 
 B&= \big(T^2 + T^6\big)e^{n_B}
\end{aligned}
$$




\begin{center}
\begin{tabular}{c}
Table 2: Common cause experiment results \\
\begin{tabular}{llrrr}
        \hline
        \multicolumn{5}{c}{\textbf{Common Cause}}\\
        Experiment & Algorithm & Normal & Uniform & Expon.  \\
        \cline{1-5}
        \multirow{ 2}{*}{1} & modKCDC & $96\%$    & $95\%$ &  $97\%$ \\
        & modIGCI & $99\%$    & $96\%$ &  $99\%$ \\
         
         \cline{1-5}
\multirow{3}{*}{2}& modKCDC & $98\%$    & $95\%$ &  $96\%$  \\
& modIGCI & $100\%$    & $100\%$ &  $100\%$ \\
 & CAN      & $80\%$   & $66\%$ &  $100\%$   \\
         \cline{1-5}
        
\multirow{2}{*}{3} & modKCDC  & $94\%$    & $99\%$ &  $95\%$  \\
& modIGCI & $98\%$    & $96\%$ &  $97\%$ \\
         \cline{1-5}
\multirow{2}{*}{4} & modKCDC    & $95\%$    & $96\%$ &  $96\%$  \\
& modIGCI & $96\%$    & $96\%$ &  $97\%$ \\
         \cline{1-5}
\multirow{2}{*}{5} & modKCDC & $97\%$    & $100\%$ &  $95\%$  \\
& modIGCI & $95\%$    & $100\%$ &  $94\%$ \\
         \cline{1-5}
\multirow{2}{*}{6} & modKCDC   & $96\%$    & $95\%$ &  $96\%$  \\
& modIGCI & $94\%$    & $96\%$ &  $93\%$ \\
\end{tabular}
\end{tabular}
\end{center}

Here, modKCDC and modIGCI achieve impressive accuracy across a range of different synthetic noise functions---including additive and multiplicative noise---and diverse distributions over these noise terms.  Despite the fact that neither KCDC nor IGCI could detect latent common causes, our method turned them into algorithms that could detect them. CAN performed well on the only experiment where it managed to fit a model, (2). Again, accuracy was only reported when it fit a model, in principle giving CAN an added advantage. In (2), this happened in 5/10, 3/10, 1/10 cases for the different noise distributions. CAN failed to fit a model in (1), (3)-(6) above, again demonstrating the difficulty in distinguishing common and directed causes.


\subsection{Common cause robustness tests}
In the above experiments we tested the robustness of our algorithms in different noise regimes. We now test robustness to more complex functional relationships. First, we test the accuracy of the algorithms in a regime beyond additive and multiplicative noise. We then test on functions drawn from Gaussian Processes. 
 Results are presented in Table~3.
\subsubsection{Complex noise}
$$ \begin{aligned}
(1) \text{ } A&= T^5 - \sin{(T^2 n_A)}, \quad (2) \text{ } A= T\sin\left(10 T |n_A|\right), \\
B&= \big(\log(T^4+10))^{2n_B} \quad\quad B= \log(T + 10) + T^{2|n_B|}
\end{aligned}
$$




\subsubsection{Gaussian Process generators}
\begin{enumerate}[(1)]
\setcounter{enumi}{2}
    \item Let $A=f(T)e^{n_A}$ \& $B=g(T)e^{n_B}$, with $f$, $g$ drawn from the same Gaussian Process whose kernel is a sum of polynomial (with only $T^2 + T^6$ terms) and periodic exponential. 
    \item $f$ is drawn from a GP with polynomial kernel (only $T^2 + T^6$ terms) and $g$ is drawn from GP with a sum of polynomial (only $T^3 + T^5$ terms) and a periodic exponential kernel.
\end{enumerate}





\begin{center}
\begin{tabular}{c}
Table 3: Complex noise common cause experiment results \\
\begin{tabular}{lllr}
\hline
        \multicolumn{3}{c}{\textbf{Complex Noise Common Cause}}\\
       Experiment & Algorithm & Result \\
    \cline{1-3}
        \multirow{2}{*}{1} & modKCDC & $93\%$ \\
        & modIGCI & $95\%$ \\
    \cline{1-3}
        \multirow{2}{*}{2} & modKCDC & $89\%$\\
        & modIGCI & $91\%$ \\
    \cline{1-3}
       \multirow{2}{*}{3} & modKCDC & $90\%$\\
       & modIGCI & $95\%$ \\
    \cline{1-3}
        \multirow{2}{*}{4} & modKCDC & $91\%$\\
        & modIGCI & $91\%$ \\
    \end{tabular}
    \end{tabular}
    \end{center} 

Both modKCDC and modIGCI perform well in these datasets beyond additive and multiplicative noise. Experiments (3) and (4) provide an analysis of robustness to changes in the parameters of the underlying functions. 


\subsection{Sensitivity to the assumption of small influence of latent noise} \label{section: sensitivity}
\begin{figure}
    \centering
    \includegraphics[scale=.5]{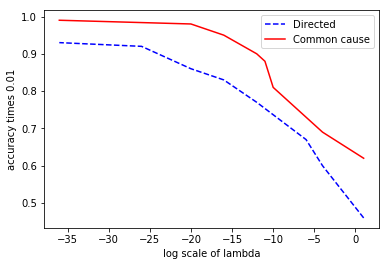}
    \caption{Effect of increasing $\lambda$ on algorithm accuracy.}
    \label{figure: assumption} 
\end{figure}

As discussed in Section~\ref{section: reduction to common causes and directed causal discovery}, a key assumption underlying our heuristic is that in the case where $A \leftarrow C \rightarrow B$ is the true causal structure, applying manifold learning to $A,B$ samples should result in an approximate parametrisation of $C$. Intuitively, this happens if the influence of noise terms is small. That is if the influence of $n_A$ in $A=f(C,n_A)$ and $n_B$ in $B=g(C,n_B)$ does not preclude $C$ from being learnt. The sensitivity of our heuristic to this assumption will now be empirically tested using modIGCI. We synthetically generate both directed and common cause data from models which violate the assumption. Directed data is generated by sampling $A\in \mathcal{N}(0,1)$ and $u_A\in \mathcal{N}(0,1)$ and inputting to
$$B= \sin{(10A)} + e^{3A} + \lambda e^{e^{n_B}}$$
Common cause data is generated by sampling $T\in \mathcal{N}(0,1)$, $u_A\in \mathcal{N}(0,1)$ \& $u_B\in \mathcal{N}(0,1)$ and inserting to
$$
\begin{aligned}
A &= \sin(3t) + \lambda e^{e^{n_A}} \\
B &= \log(T+10) + \lambda e^{e^{n_B}}
\end{aligned} 
$$
Here, the latent noise terms influence the observed variables via a double exponential function, which violates the assumption underlying Intuition Aid~\ref{lemma 3}.
In both cases, the smaller $\lambda$ is, the smaller the influence of the noise terms. We would expect that the smaller $\lambda$, the higher the accuracy of our method, and the larger $\lambda$, the lower the accuracy of our method. This intuition is borne out in the experiments. Results are plotted in Figure~\ref{figure: assumption}. 

Despite the fact that the noise influences the observed variables via a double exponential function, the accuracy of modIGCI is quite resilient. For instance, when $\lambda=1$ the accuracy on directed data is 46\% and the accuracy on common cause data is 62\%. Full results are in Figure~\ref{figure: assumption}. This gives hope that identifiability can be demonstrated in a wider array of settings than was shown in Section~\ref{section: reduction to common causes and directed causal discovery}.

\subsection{Real data}  
\subsubsection{Directed data}

We now test on real world data. For this we consider the T{\"u}bingen 
cause-effect pairs dataset,\footnote{https://webdav.tuebingen.mpg.de/cause-effect/} discussed in \citep{mooij2016distinguishing}. This dataset is comprised of real-world cause-effect samples that are collected across diverse subject areas, from physics to education, with the true causal direction provided by human domain experts. 

We use T{\"u}bingen cause-effect pairs dataset version 1.0 which consists of 100 cause-effect pairs. Each pair is assigned a weight in order to account for potential sources of bias given that different pairs are sometimes selected from the same multivariable dataset. Following the wide-spread approach present in the literature of testing only on scalar-valued pairs, we remove the multivariate pairs 52, 53, 54, 55 and 71 from T{\"u}bingen cause-effect pairs dataset to ensure a fair comparison to previous work. 

For modKCDC and modIGCI, we use the thresholds fixed in Section~\ref{section:experiments}, which are reported in the Appendix. We tested against KCDC and IGCI and also against the additive noise model (ANM) algorithm of \citep{Noise}, a well known directed causal discovery algorithm discussed in Section~\ref{Previous approaches to causal discovery}. The results are shown in Table 4.

\begin{center}
\begin{tabular}{c}
Table 4: Cause-effect pairs real data results \\
\begin{tabular}{lr} 
\hline
\multicolumn{2}{c}{\textbf{T{\"u}bingen cause-effect pairs dataset}}\\
        Algorithm & Result \\
        \cline{1-2}
         KCDC & $71\%$ \\
         modKCDC & $69.6\%$ \\
         IGCI & $72\%$\\
         modIGCI & $67.4\%$ \\
        ANM & $59.5\%$\\
    \end{tabular}
    \end{tabular}
    \end{center}
    
It is important to note is that both modKCDC and modIGCI achieved largely similar performance to that of KCDC and IGCI, illustrating that our method does not substantially reduce the ability of the original, unmodified algorithms to detect direct causal relations. The performance of KCDC is slightly worse than the performance reported in \citep{mitrovic2018causal}, possibly due to the specific hyperparameter optimisation undertaken by those authors.  Note that an adaptation of the KCDC approach, with multiple kernel majority vote for each cause-effect pair, is also undertaken in \citep{mitrovic2018causal} with better accuracies than KCDC with fixed hyperparameters. We did not implement the majority-vote approach here. The performance of IGCI and ANM is also worse than reported in \citep{mooij2016distinguishing}, again possibly because of the specific extensive hyperparameter optimisation undertaken by those authors. However, the performance of IGCI and ANM reported here is largely similar to that reported in \citep{mitrovic2018causal}. 

It is instructive to consider \emph{NLSchools data}, the $99$th pair in the T{\"u}bingen 
cause-effect data. Here `socioeconomic status' is expected to be a direct cause of `language score'. However, some researchers have argued that the correlation between these two attributes could be genetic in nature, implying the underlying structure is purely common cause. When applied on this data, modKCDC returns $\text{mean}(\{\Delta\})=0.933$ and $\text{var}(\{\Delta\})=0.007$. 
modKCDC's output---with its high mean and low variance---is within the thresholds fixed in Section~\ref{section:experiments}, which are reported in the Appendix, and is hence indicative of a directed causal structure. 
Hence, 
modKCDC 
returns a direct cause from `socioeconomic status' to `language score', ruling out a purely common cause explanation of this data and agreeing with the ground truth. A plot of this data is given in Fig.~\ref{fig:real_data} (a).

\subsubsection{Common cause data}

\begin{figure}[t] 
\centering
\begin{subfigure}
\centering
(a)\includegraphics[scale=0.22]{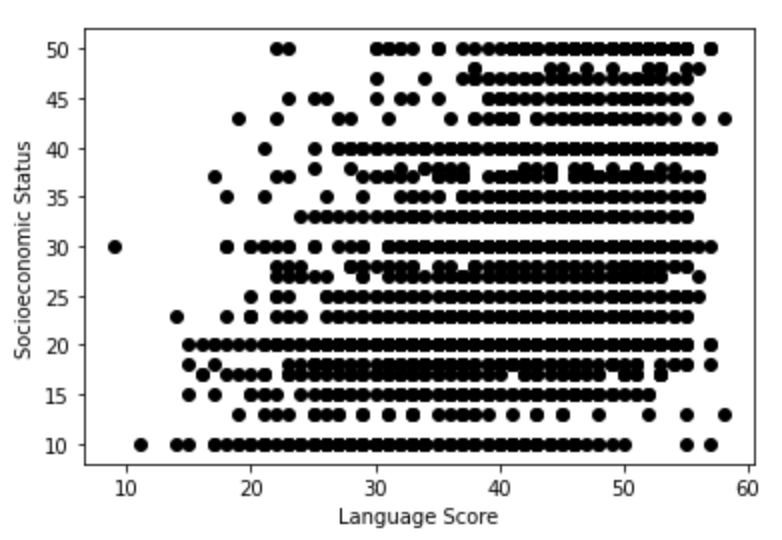}
\end{subfigure}
\quad
\begin{subfigure}
\centering
(b) \includegraphics[scale=0.22]{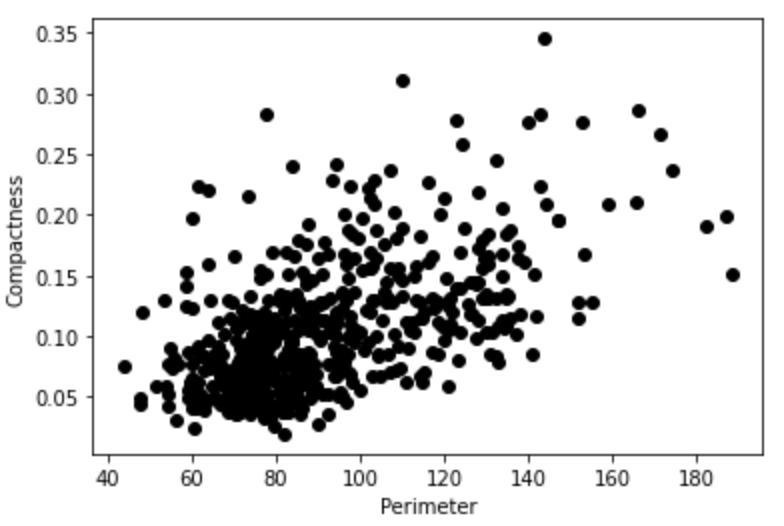}
\end{subfigure}
\quad
\begin{subfigure}
\centering
(c) \includegraphics[scale=0.22]{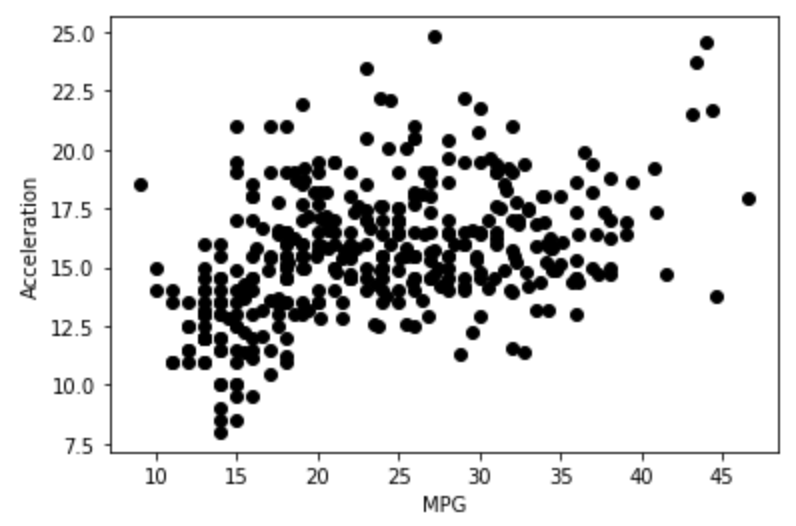}
\end{subfigure}
\caption{
Plots of real world data.
}
\label{fig:real_data}
\end{figure}

Next, we consider \emph{Breast Cancer Wisconsin (Diagnostic) Data} \citep{UCI}. Here the tumour (malignant or benign) is the common cause of two attributes, `perimeter' and `compactness', as they are conditionally independent given it. When run on this `perimeter' and `compactness' data, modKCDC returns $\text{mean}(\{\Delta\})=0.491$ and $\text{var}(\{\Delta\})=0.081$, and modIGCI returns $\text{mean}(\{\Delta\})=0.308$ and $\text{var}(\{\Delta\})=0.062$. In both cases, $\text{mean}(\{\Delta\})$ is relatively small, with $\text{var}(\{\Delta\})$ in both cases relatively large. From the discussion in Section~\ref{section: decision criteria}, we should conclude that both algorithms are consistent with a common cause.
In addition, all values are within the thresholds fixed in Section~\ref{section:experiments}, which are reported in the Appendix, for a common cause.
We conclude that both algorithms return a common cause between `perimeter' \& `compactness', agreeing with the ground truth. A plot of this data is given in Fig.~\ref{fig:real_data} (b).

Next we consider the \emph{AutoMPG} data set from \citep{UCI}. Here, attributes `acceleration' and `MPG' (miles per gallon) are correlated but have a common cause, the model year. 
This is checked by performing a conditional independence test between `acceleration' and `MPG' given model year. When applied on this data, modKCDC returns $\text{mean}(\{\Delta\})=0.639$ and $\text{var}(\{\Delta\})=0.086$, with modIGCI yielding $\text{mean}(\{\Delta\})=0.124$ and $\text{var}(\{\Delta\})=0.00062$. As per the thresholds reported in the Appendix, we conclude that modKCDC returns a common cause between `acceleration' and `MPG'. While modIGCI returned a low $\text{mean}(\{\Delta\})$ value---which is indicative of a common cause---its value for $\text{var}(\{\Delta\})$ is also low, which is not in general consistent with one. Given this relatively large size difference, following the discussion at the end of Section~\ref{section: decision criteria} one could conclude that modIGCI's output is consistent with a common cause  between `acceleration' and `MPG'. A plot is given in Fig.~\ref{fig:real_data} (c).


\section{Conclusion} \label{section: conclusion}

We devised a method to turn a purely directed causal discovery algorithm into one that can also detect latent common causes. In our experiments we took KCDC \citep{mitrovic2018causal} and IGCI \citep{daniusis2012inferring}, which each could distinguish directed causes but could not detect latent common causes, and showed our method enabled detection of latent common causes while preserving accuracy in distinguishing directed causes. We also discussed some open problems regarding the identifiability of our algorithm, which we hope will stimulate future research.

The setting of the thresholds for the decision criterion from Section~\ref{section: decision criteria} is a challenging problem. The use of domain heuristic employed here follows both the approach of CAN \citep{Confounders}, described in Section~\ref{section: CAN model}, and KCDC \citep{mitrovic2018causal}, outlined just before Definition~\ref{def: under}. Despite its heuristic nature, it worked quite well in our experiments, despite how varied they were. Ongoing research will investigate the use of statistical tests to analyse the thresholds. 
 
In future work, applications to the burgeoning field of quantum causal models \citep{allen2017quantum,lee2018towards,lee2019device,wolfe2016inflation,chaves2015information} will be explored.

\bibliographystyle{unsrtnat}

\bibliography{references} 

\appendix 

\section{Decision thresholds for experimental section of main paper}

In Section~4 of the main paper, we use synthetic directed data to set the thresholds for the decision criteria---discussed in Section~3.2 of the main paper---of our modifications of KCDC \citep{mitrovic2018causal} and IGCI \citep{daniusis2012inferring}, termed modKCDC and modIGCI  respectively. To set the thresholds, we generated synthetic data from directed causal models with additive, multiplicative, and complex noise functions---similar to the ones used in the Experiments section---and determined the threshold values which approximately reproduced the performance of the original directed causal discovery algorithms on these three different models. These thresholds are then used in all experiments. While this is somewhat ad-hoc, it is similar to the way CAN's thresholds were set in \citep{janzing2012information}, and worked well in our experiments despite the fact that they were quite diverse in nature. The thresholds were set following the bootstrapping approach detailed in Section~3.2, using 25 bootstraps each randomly sampling 95\% of the data. 

The thresholds we set for modKCDC are as follows:
\begin{enumerate}
    \item If $\text{mean}(\{\Delta\})>0.9$ and $\text{var}(\{\Delta\})\leq 0.06$, then we output a directed causal structure oriented using the original causal discovery algorithm. 
    \begin{itemize}
    \item The regime $\text{mean}(\{\Delta\})>0.9$ and $\text{var}(\{\Delta\})> 0.06$ is a failure mode of our algorithm. 
    \end{itemize}
    \item If $\text{mean}(\{\Delta\})< 0.25$ and $\text{var}(\{\Delta\})>0.03$, a common cause structure is output. 
    \begin{itemize}
    \item The regime $\text{mean}(\{\Delta\})< 0.25$ and $\text{var}(\{\Delta\}) < 0.03$ is a failure mode of our algorithm. 
    \end{itemize}
    \item If $0.65<\text{mean}(\{\Delta\})\leq 0.9$ and $\text{var}(\{\Delta\})\leq 0.06$, then a directed causal structure is output. 
    \item But if $0.65<\text{mean}(\{\Delta\})\leq 0.9$ and $\text{var}(\{\Delta\})>0.06$, then a common cause structure is output. 
    \item If $0.25\leq \text{mean}(\{\Delta\})\leq 0.65$ and $\text{var}(\{\Delta\})\leq 0.03$, then a directed causal structure is output. 
    \item But if $0.25\leq \text{mean}(\{\Delta\})\leq 0.65$ and $\text{var}(\{\Delta\})>0.03$, then a common cause structure is output.
    \end{enumerate}

\noindent The thresholds we set for modIGCI are as follows:
\begin{enumerate}
    \item If $\text{mean}(\{\Delta\})>0.9$ and $\text{var}(\{\Delta\})\leq 0.02$, then we output a directed causal structure oriented using the original causal discovery algorithm.
    \begin{itemize}
    \item The regime $\text{mean}(\{\Delta\})>0.9$ and $\text{var}(\{\Delta\})> 0.02$ is a failure mode of our algorithm.
    \end{itemize}
    \item If $\text{mean}(\{\Delta\})< 0.25$ and $\text{var}(\{\Delta\})>0.01$, a common cause structure is output
    \begin{itemize}
        \item The regime $\text{mean}(\{\Delta\})< 0.25$ and $\text{var}(\{\Delta\}) \leq 0.01$ is a failure mode of our algorithm.
    \end{itemize} 
    \item If $0.45<\text{mean}(\{\Delta\})\leq 0.9$ and $\text{var}(\{\Delta\})\leq 0.02$, then a directed causal structure is output. 
    \item But if $0.45<\text{mean}(\{\Delta\})\leq 0.9$ and $\text{var}(\{\Delta\})>0.02$, then a common cause structure is output.
    \item If $0.25\leq \text{mean}(\{\Delta\})\leq  0.45$ and $\text{var}(\{\Delta\})\leq 0.01$, then a directed causal structure is output. 
    \item But if $0.25\leq \text{mean}(\{\Delta\})\leq 0.45$ and $\text{var}(\{\Delta\})>0.01$, then a common cause structure is output.
\end{enumerate}

The thresholds set for CAN, described in Section~2.1 of the main paper, are as follows. Let $r =\text{var}(u_A)/\text{var}(u_B)$. If $r \leq 0.65$ then a $A\rightarrow B$ is returned. If $r \geq 1.65$, then $A \leftarrow B$ is returned. If $0.65 < r < 1.65$, then a common cause structure is returned.

\end{document}